\documentclass[10pt,twocolumn,letterpaper]{article}

\usepackage{cvpr}              

\usepackage{graphicx}
\usepackage{amsthm}
\usepackage{amsmath}
\usepackage{amssymb}
\usepackage{booktabs}
\usepackage[ruled,linesnumbered]{algorithm2e}
\usepackage{setspace}
\usepackage{tabularx}
\usepackage{dsfont}

\usepackage[pagebackref,breaklinks,colorlinks]{hyperref}

\usepackage[capitalize]{cleveref}
\crefname{section}{Sec.}{Secs.}
\Crefname{section}{Section}{Sections}
\Crefname{table}{Table}{Tables}
\crefname{table}{Tab.}{Tabs.}

\def\cvprPaperID{3619} 
\def\confName{CVPR}
\def\confYear{2022}

\begin{document}

\title{Source-Free Domain Adaptation via Distribution Estimation}
\author{
Ning Ding$^{1}$, Yixing Xu$^{2}$, Yehui Tang$^{1,2}$, Chao Xu$^{1}$, Yunhe Wang$^{2*}$, Dacheng Tao$^{3}$\\
\small $^{1\,}$Key Lab of Machine Perception (MOE), School of Artificial Intelligence, Peking University\\
\small $^{2\,}$Huawei Noah’s Ark Lab\qquad$^{3\,}$JD Explore Academy, China\\
\tt\small dingning@stu.pku.edu.cn, \{yixing.xu, yunhe.wang\}@huawei.com, yhtang@pku.edu.cn\\
\tt\small  xuchao@cis.pku.edu.cn, dacheng.tao@gmail.com
}

\maketitle
\let\thefootnote\relax\footnotetext{* Corresponding author.}
\begin{abstract}
Domain Adaptation aims to transfer the knowledge learned from a labeled source domain to an unlabeled target domain whose data distributions are different. However, the training data in source domain required by most of the existing methods is usually unavailable in real-world applications due to privacy preserving policies. Recently, Source-Free Domain Adaptation (SFDA) has drawn much attention, which tries to tackle domain adaptation problem without using source data. In this work, we propose a novel framework called SFDA-DE to address SFDA task via source \textbf{D}istribution \textbf{E}stimation. Firstly, we produce robust pseudo-labels for target data with spherical k-means clustering, whose initial class centers are the weight vectors (anchors) learned by the classifier of pretrained model. Furthermore, we propose to estimate
the class-conditioned feature distribution of source domain by exploiting target data and corresponding anchors. Finally, we sample surrogate features from the estimated distribution, which are then utilized to align two domains by minimizing a contrastive adaptation loss function. Extensive experiments show that the proposed method achieves state-of-the-art performance on multiple DA benchmarks, and even outperforms traditional DA methods which require plenty of source data.
\end{abstract}

\begin{figure}[t]
\centering
\includegraphics[width=1.\linewidth]{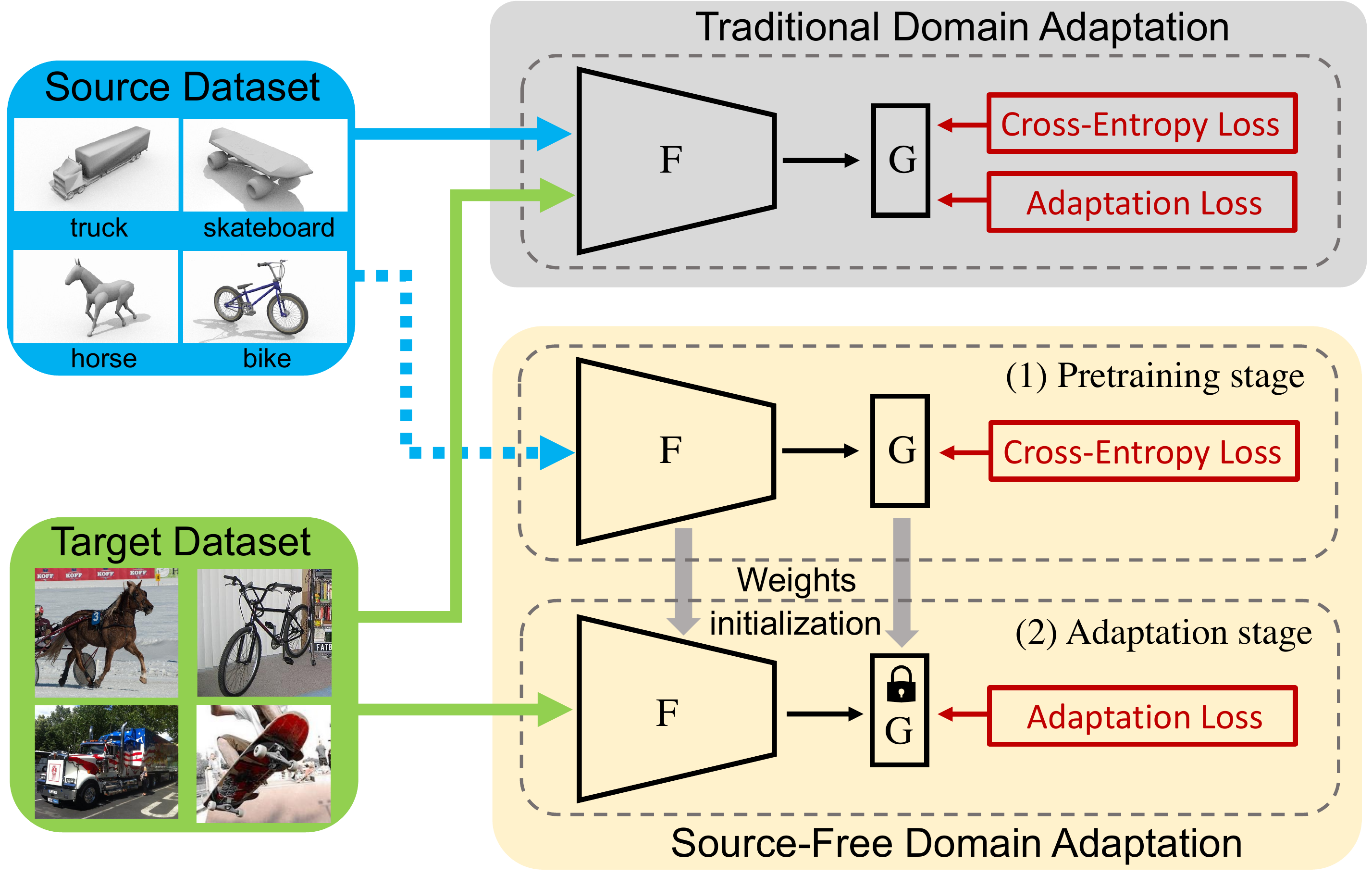}
\vspace{-0.6cm}
\caption{(Top) Traditional domain adaptation methods require data from both source domain and target domain simultaneously. (Bottom) In source-free domain adaptation, source data can only be used in the pretraining stage, and cannot be accessed in the later adaptation stage. Adaptation is achieved by utilizing target data and the model pretrained on source domain.}
\label{fig:teaser}
\end{figure}

\section{Introduction}
\label{sec:intro}
In the past few years, deep Convolutional Neural Networks (CNNs) have achieved remarkable performance on many visual tasks such as classification \cite{krizhevsky2012imagenet}, object detection \cite{girshick2015fast}, semantic segmentation \cite{long2015fully}, \textit{etc.} However, the success of CNNs relies heavily on the hypothesis that the distributions of the training data is identical to that of the test data. Thus, models trained with data from a certain scenario (source domain) can hardly generalize well to other real-world application scenarios (target domains), and may suffer from severe performance drop. Moreover, the difficulty of collecting enough labeled training data also hinders CNNs from directly learning with target domain data. Unfortunately, CNNs deployed in real-world scenarios always encounter new situations, such as the change of weather and variation of illumination in autonomous driving.

Therefore, a lot of attention is paid to the domain shift problem~\cite{sugiyama2007direct, ben2007analysis} mentioned above, and \textit{Domain Adaptation} (DA) theory has been developed to solve it. DA algorithms directly help deep models transfer knowledge learned from a fully annotated source domain to a separately distributed target domain whose annotations are entirely unavailable. Existing advances in deep learning-based DA methods~\cite{long2018conditional,GVB,li2021transferable,FixBi} generally achieve model transferability by means of mapping two different data distributions simultaneously into a mutual feature space shared across two domains.

However, people are getting more aware of the importance of privacy data protection nowadays. Strict policies regarding data privacy concerns have been published all around the world. More AI companies also choose to open source their pretrained models only, yet keep the source dataset used for training unreleased~\cite{sun2017revisiting}. Therefore, most of the traditional DA methods become infeasible to transfer knowledge to target domain when source data is no longer accessible since these methods basically assume that data from both source domain and target domain is available.

To overcome this data-absent problem, some recent works~\cite{SHOT, 3CGAN, universalSFDA, kim2021domain,xu2019positive} explored more general approaches to achieve domain adaptation without accessing source data. Only unlabeled target domain data and the model pretrained on source domain are required to accomplish the cross-domain knowledge transfer. Such a new unsupervised learning setting for domain adaptation task is called \textit{Source-Free Domain Adaptation} (SFDA). 
SHOT~\cite{SHOT} utilizes information maximization and entropy minimization. 3C-GAN~\cite{3CGAN} uses a generative model to enrich target data to enhances model performance. G-SFDA~\cite{G-SFDA} learns different feature activations by exploiting neighborhood structure of target data. A$^2$Net~\cite{A2Net} introduces a new classifier and adopt adversarial training to align two domains. 
Despite the fact that these SFDA methods utilize the source domain knowledge contained by the pretrained model, none of them explicitly align the distributions between source domain and target domain to achieve adaptation.

In this paper, we focus on image classification task under SFDA setting. We manage to estimate the source distribution without accessing source data. Specifically, we utilize the domain information captured by the model pretrained on source data and treat the weights learned by source classifier as class anchors. Then, these anchors are used as the initialization of feature center for each class and spherical k-means is performed to cluster target features in order to produce robust pseudo-labels for target data. Furthermore, we dynamically estimate the feature distributions of source domain class-wisely by utilizing the semantic statistics of target data along with their corresponding anchors, which is called Source Distributions Estimation (\textbf{SDE}). Finally, we sample surrogate features from distributions derived from SDE to simulate the real but unknown source features, and then align them with target features by minimizing a contrastive adaptation loss function to facilitate source-free domain adaptation. In short, if the feature distribution of target domain is well-aligned with source domain, the source classifier will naturally adapt to the target domain data.

We validate our proposed SFDA-DE method on three public DA benchmarks: Office-31~\cite{office31}, Office-Home~\cite{venkateswara2017deep} and VisDA-2017~\cite{peng2017visda}. Experiment results show that the proposed SFDA-DE method achieves state-of-the-art performance on Office-Home (72.9\%) and VisDA-2017 (86.5\%) among all SFDA methods, and is even superior to some recently proposed traditional DA methods that require accessing source domain data.

\section{Related Work}
\paragraph{Traditional domain adaptation.}
Domain Adaptation (DA) as a research topic has been studied for a long time \cite{ben2007analysis}. With the emergence of deep learning \cite{krizhevsky2012imagenet, simonyan2014very}, CNNs with superior capacity to capture high level features become the first choice to perform adaptation. As a result, many related tasks have been developed in the field of visual DA, such as multi-source DA \cite{xu2018deep,peng2019moment}, semi-supervised DA \cite{he2020classification, saito2019semi}, partial DA \cite{cao2018partial}, open set DA \cite{panareda2017open, compounddomainadaptation}, universal DA \cite{saito2020universal}, \etc DA aims to improve the generalizability of a model which is learned on a labeled source domain. When fed with data drawn from a different target distribution, model performance declines drastically. This is referred to as \textit{covariate shift} \cite{song2009hilbert, sugiyama2007direct, sugiyama2008direct} or \textit{domain shift} problem. To tackle this problem, lots of methods try to align feature distributions of different domains via minimizing Maximum Mean Discrepancy (MMD)~\cite{long2015learning, long2016unsupervised, long2017deep, sugiyama2007direct}, which is a non-parametric kernel function embedded into reproducing kernel Hilbert space (RKHS) to measure the difference between two probability distributions \cite{gretton2012kernel,iyer2014maximum}. Moreover, Kang \etal~\cite{CAN} incorporates contrastive learning technique \cite{chen2020simple} into MMD-based method to further boost model transferability. Meanwhile, Zellinger \etal~\cite{zellinger2017central} and Sun \etal~\cite{sun2017correlation} propose to align high order statistics captured by networks like central moment to achieve domain adaptation.
Apart from directly aligning two distributions, some recent works~\cite{ganin2016domain,tzeng2017adversarial,long2018conditional} employ adversarial training by adding an extra feature discriminator. In this way, the networks are forced to learn domain-invariant features to confuse the discriminator.

\begin{figure*}[t]
  \centering
   \includegraphics[width=0.75\linewidth]{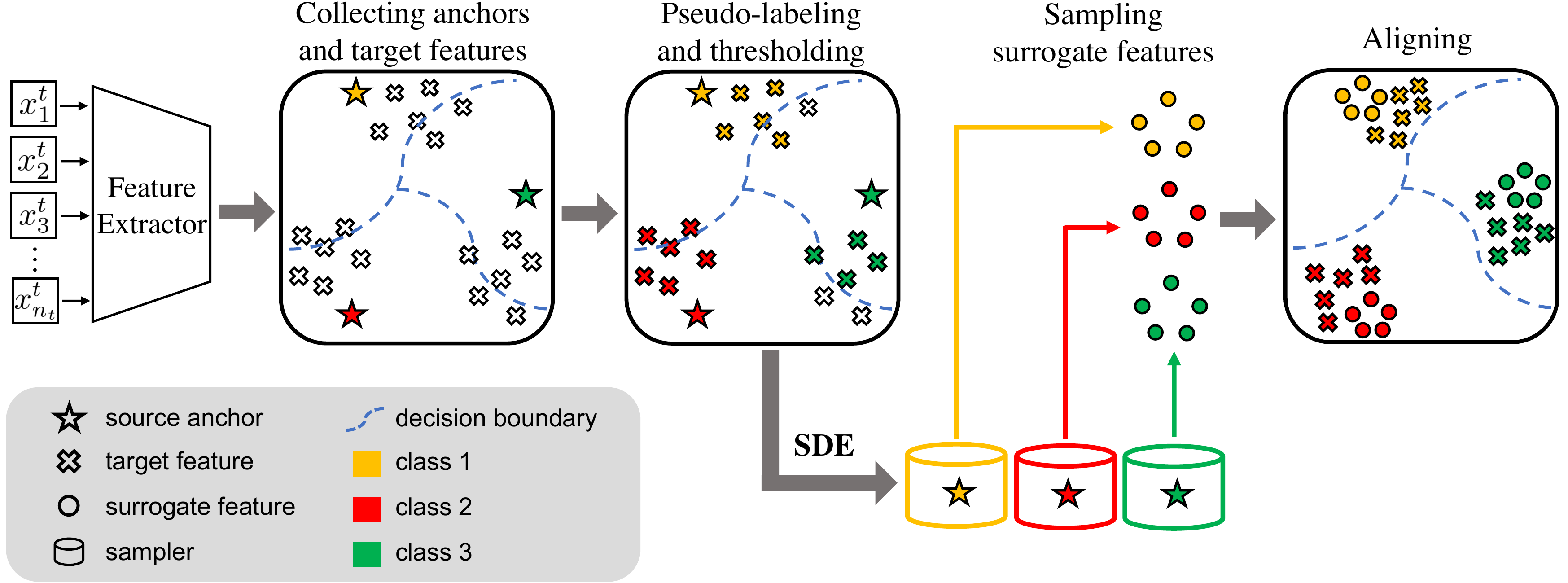}
    \vspace{-0.25cm}
   \caption{Overall training pipeline of our proposed  SFDA-DE method.}
   \label{fig:onecol}
\vspace{-0.4cm}
\end{figure*}

\vspace{-0.5cm}
\paragraph{Source-free domain adaptation.}
All the methods mentioned above expect both labeled source data and unlabeled target data to achieve domain adaptation process, which is often impractical in real-world scenario. In most cases, one can only access the unlabeled target data and the model pretrained by source data. To this end, some recent works~\cite{SHOT,3CGAN,liu2021source,A2Net,G-SFDA,universalSFDA,VDM-DA,xu2021learning} regarding source-free domain adaptation emerge. These methods provide solutions to adapt the model to unseen domains without using original training data. SHOT~\cite{SHOT} utilizes information maximization and entropy minimization via pseudo-labeling strategy to adapt the trained classifier to target features. \cite{3CGAN,liu2021source} both use generative models to model the distribution of target data by generating target-style images to enhance the model performance on target domain. G-SFDA~\cite{G-SFDA} forces the network to activate different channels for different domains while paying attention to the neighborhood structure of data. A$^2$Net~\cite{A2Net} introduces a new target classifier to align two domains via adversarial training manner. SoFA~\cite{yeh2021sofa} uses a Variational Auto-Encoder to encode target distribution in latent space while reconstructing the target data in image space to constrain the latent features. Many of the above methods freeze the source classifier during adaptation to preserve class information, and assign pseudo-labels based on the classifier's output. Here we follow the idea of freezing the source classifier~\cite{SHOT} but use a more robust pseudo-labeling strategy via spherical k-means clustering. Moreover, we propose Source Distribution Estimation (SDE), aiming to approximate the source feature distribution without accessing the source data. After that, the target distribution can be directly aligned with the estimated distribution to adapt to the source classifier.

\vspace{-0.2cm}
\section{Method}
\vspace{-0.2cm}
In this section, we first describe the problem setting for source-free domain adaptation and notations to be used afterward. Then we elaborate our proposed  SFDA-DE method in three steps to address SFDA problem. First of all, we obtain robust pseudo-labels for target data by utilizing source anchors and spherical k-means clustering. Secondly, we estimate the class-conditioned feature distribution of source domain. Finally, surrogate features are sampled from the estimated distribution to align two domains by minimizing a contrastive adaptation loss function.

\vspace{-0.2cm}
\subsection{Preliminaries and notations}
\vspace{-0.2cm}
In this paper, we use $\mathcal{D}_s=\{(x_i^s,y_i^s )\}_{i=1}^{n_s}$ to denote the source domain dataset with $n_s$ labeled samples, where $y\in\mathcal Y\subseteq\mathbb R^K $ is the one-hot ground-truth label and $K$ is the total number of classes of the label set $\mathcal{C}=\{1,2,\cdots,K\}$. $\mathcal{D}_t=\{(x_i^t)\}_{i=1}^{n_t}$ denotes the target domain dataset with $n_t$ unlabeled samples which has the same underlying label set $\mathcal{C}$ as that of $\mathcal{D}_s$. In SFDA scenario, we have access to the model $\mathbf{G}(\mathbf{F}(\cdot))$ which is already pretrained on $\mathcal{D}_s$ in a supervised manner by cross-entropy loss, where $\mathbf{F}$ denotes the CNN feature extractor followed by a linear classifier $\mathbf{G}$. During training, only data in $\mathcal{D}_t$ is available and no data in $\mathcal{D}_s$ can be used. Besides, we use $f=\mathbf{F}(x)\in \mathds{R}^{m}$ to denote $m$-dimensional feature representations and use $\mathbf{w}^G\in \mathds{R}^{m\times K}$ to denote the
weights learned by $\mathbf{G}$, where $\mathbf{w}_k^G\in \mathds{R}^{m}$ is the $k$-th weight vector of $\mathbf{w}^G$.

\subsection{Pseudo-labeling by exploiting anchors}
\label{sec:pseudolabel}
\vspace{-0.2cm}
In many works, pseudo-labeling is an important technique to obtain category information for those unlabeled samples and is usually realized by exploiting the highly-confident outputs derived by the classifier. However in SFDA task, the classifier $\mathbf{G}$ is pretrained on source domain data and will encounter the distribution shift problem when classifying target domain data. Therefore, it’s crucial to find a robust way to solve the distribution shift problem and assign correct labels to unlabeled target data. Thus, we consider obtaining pseudo-labeling via spherical k-means.

Specifically, given a class label predicted by the linear classifier $\mathbf{G}$ as
\vspace{-0.2cm}
\begin{equation}
\hat{y}_i = \mathop{\arg\max}_{k} f_i^\top \mathbf{w}_k^G ~,~ k\in\mathcal{C}=\{1,2,\cdots,K\}~,
\label{eq_cls}
\vspace{-0.2cm}
\end{equation}
where $\hat{y}_i\in\mathds{R}^K$ is the logits vector before softmax. Note that each element in the class probability vector is derived by the dot product between the feature and each weight vector of the classifier. Thus, data of the $k$-th class tends to yield feature representation that activates the $k$-th weight vector in $\mathbf{G}$. Features of data from the $k$-th class should gather around $\mathbf{w}_k^G$. Therefore, $\mathbf{w}_k^G$ can be treated as an \textbf{anchor} of the $k$-th class which contains overall characteristics that represent the whole $k$-th class.

In SFDA task, target features would drift away from source anchors which makes it hard to directly predict labels for target data with pretrained classifier $\mathbf{G}$. Thus, we propose to assign pseudo-labels for target data via spherical k-means. We first cluster the target data by setting anchors as the initial cluster centers: $\mathcal{A}_k^{(0)} = \mathbf{w}_k^G$. Then we perform spherical k-means iteratively between (1) assigning pseudo-labels via minimum-distance classifier: $\hat{y}_i^t = \mathop{\arg\min}_{k} Dist(\mathcal{A}_k^{(m)}, f_i^t)$ and (2) computing new cluster centers $\mathcal{A}_k^{(m+1)} = \frac{\sum_{i=1}^{n_t} \mathds{1}(\hat{y}_i^t=k) f_i^t}{\sum_{i=1}^{n_t} \mathds{1}(\hat{y}_i^t=k)} $, where $Dist(\mathbf{a},\mathbf{b})=\frac{1}{2}(1-\frac{\mathbf{a}^\top \mathbf{b}}{|\mathbf{a}|\cdot|\mathbf{b}|})$ is the cosine distance, $m$ denotes the number of current iterations and $\mathds{1}(\cdot)$ is the indicator function. Iteration will stop when all class centers converge. After clustering is done, a confidence threshold $\tau\in (0,1)$ is set to filter out ambiguous samples so that a confidently pseudo-labeled target dataset $\mathcal{D}_t'$ is constructed:
\vspace{-0.2cm}
\begin{equation}
\mathcal{D}_t'=\{(x_i^t,\hat{y}_i^t )~|~ Dist(f_i^t,\mathcal{A}_{\hat{y}_i^t})<\tau,~\hat{y}_i^t\in \mathcal{C} \}_{i=1}^{n_t'}.
\label{confidentpseido}
\vspace{-0.2cm}
\end{equation}

Given the robust pseudo-labels derived above, we use $x_{i,k}^t$ to denote the target data $x_{i}^t$ with pseudo-label $\hat{y}_{i}^t=k$, and use $f_{i,k}^t = \mathbf{F}(x_{i,k}^t)$ to denote its corresponding feature representation in the following of this paper. Similar to the idea proposed in~\cite{SHOT}, we freeze $\mathbf{G}$ to fix the source anchors in order to stabilize the adaptation to target domain.


\subsection{Source Distribution Estimation}
\label{sec:SSDE}
In traditional DA setting, feature distributions of data from both source and target domains can be estimated by mini-batch sampling from $\mathcal{D}_s$ and $\mathcal{D}_t$,  respectively. Then the target distribution can be explicitly aligned with the source one and classified by the pretrained source classifier $\mathbf{G}$~\cite{long2016unsupervised, long2017deep,CAN}. However, source data is unavailable in SFDA setting, which makes it impossible to know the source distribution. To tackle this problem, Yang \etal~\cite{G-SFDA} focuses on neighborhood structure and channel activation. Liang \etal\cite{SHOT} exploits information maximization and self-supervision to implicitly align feature representations. Nevertheless, none of the existing methods address SFDA problem by explicitly aligning the source distribution with the target distribution, and thus achieve sub-optimal results. We manage to explicitly estimate the source feature distribution without accessing source data by presenting Source Distribution Estimation (SDE) method.

Concretely, we assume feature representations of source domain follow a class-conditioned multivariate Gaussian distribution $f_{i,k}^s \sim \mathcal{N}_k^s(\mu_k^s,\Sigma_k^s)$, where $f_{i,k}^s=\mathbf{F}(x_i^s|y_i^s$$=$$k)$ and $k$$\in$$ \mathcal{C}$$=$$\{1,2,\cdots,K\}$. Essentially, $\mu_k^s$ can be viewed as the center of feature representations of the $k$-th class data in source domain and $\Sigma_k^s$ is the covariance matrix which captures the variation in features of the $k$-th class and contains rich semantic information~\cite{wang2019implicit}. Then we can use a surrogate distribution $\mathcal{N}_k^{sur}(\hat{\mu}_k^s,\hat{\Sigma}_k^s)$ to approximate the actual but unknown source distribution $\mathcal{N}_k^s$ for each class $k\in\mathcal{C}$.


A good estimator for $\mu_k^s$ should be discriminative enough and reflect the intrinsic characteristics of the $k$-th class data in source domain. If we directly use the feature mean of target data of the $k$-th class $\bar{f_k^t}=\frac{\sum_{i}f_{i,k}^t}{\sum_{x_i^t \in \mathcal{D}_t'}\mathds{1}(\hat{y}_i^t=k)}$ as an estimator for $\mu_k^s$, obviously the above conditions cannot be satisfied due to the existence of domain shift problem. Recall the observation in \cref{sec:pseudolabel} that anchors contain overall characteristics of the corresponding class. Thus, we propose to utilize anchors to calibrate the estimator for mean of the surrogate source distribution:
\vspace{-0.2cm}
\begin{equation}
\hat{\mu}_k^s = {\|\bar{f_k^t}\|}_2\cdot \frac{\mathbf{w}_k^G}{{\|\mathbf{w}_k^G\|}_2}~,~ k\in \mathcal{C},
\label{muestim}
\vspace{-0.2cm}
\end{equation}
which implies that the direction of estimated source feature mean is the same as the corresponding anchor but the scale of it is derived from target features. Another reason for the calibration is that there is usually a difference in norm between anchors and features, which is ${\|\mathbf{w}_k^G\|}_2 < {\|f_{i,k}^t\|}_2  \approx {\|f_{i,k}^s\|}_2$, empirically. Therefore, it's not appropriate to directly use anchors as the estimator of mean either.

As for covariance matrices, many works~\cite{li2021transferable,wang2019implicit,cui2014flowing,li2018semi} study the statistics of deep features and reveal that class-conditioned covariance implies the activated semantic directions and correlations between different feature channels. We assume that the intra-class semantic information of target features is roughly consistent with that of the source. Hence we derive the estimator for source covariance $\Sigma_k^s$ from statistics of target features:
\vspace{-0.2cm}
\begin{equation}
\hat{\Sigma}_k^s = \gamma\cdot {\Sigma}_k^t = \gamma\cdot \frac{\textbf{f}_k^{\,t}\cdot{\textbf{f}_k^{\,t}}^\top}{\sum\limits_{x_i^t \in \mathcal{D}_t'}\mathds{1}(\hat{y}_i^t=k)},
\label{sigmaestim}
\vspace{-0.2cm}
\end{equation}
where $\textbf{f}_k^{\,t}=[f_{1,k}^t-\bar{f_k^t},\cdots,f_{i,k}^t-\bar{f_k^t},\cdots]$ is a matrix whose columns are centralized target features of the $k$-th class in $\mathcal{D}_t'$. We use a controlling coefficient $\gamma$ to adjust the sampling range and semantic diversity of sampled surrogate features. Details of selecting $\gamma$ will be studied in \cref{sec:abl}.

By exploiting anchors and target features, we derive $K$ class-conditioned surrogate source distributions
\vspace{-0.2cm}
\begin{equation}
\mathcal{N}_k^{sur}({\|\bar{f_k^t}\|}_2\frac{\mathbf{w}_k^G}{{\|\mathbf{w}_k^G\|}_2},  \frac{\gamma\cdot\textbf{f}_k^{\,t}\cdot{\textbf{f}_k^{\,t}}^\top}{\sum\limits_{x_i^t \in \mathcal{D}_t'}\mathds{1}(\hat{y}_i^t=k)}), k\in\mathcal{C},
\label{sur_dis}
\vspace{-0.2cm}
\end{equation}
from which we can sample surrogate features $f_k^{sur} \sim\mathcal{N}_k^{sur}(\hat{\mu}_k^s,\hat{\Sigma}_k^s)$ to simulate the real source features.

\subsection{Source-free domain adaptation}
In the previous section, we are able to estimate the source distribution without accessing source data by exploiting domain knowledge preserved in the pretrained model with the proposed SDE method. Thus, we can sample data from the estimated distribution as surrogate source data, and the SFDA problem becomes the traditional DA problem. We adopt Contrastive Domain Discrepancy (CDD) introduced by Kang \etal~\cite{CAN} to explicitly align the target distribution with the estimated source distribution.

Specifically, we choose a random subset $\mathcal{C}'\subset\mathcal{C}$ from the label set $\mathcal{C}$$=$$\{1,2,\cdots,K\}$ before each forward pass. Then for each $k\in\mathcal{C}'$, we sample $n_b$ target images from $\mathcal{D}_t'$ to construct a set of data $\{\{(x_i^t, \hat{y}_i^t=k)\}_{i=1}^{n_b}| k\in\mathcal{C}'\}$ and derive the target mini-batch $\{\{f_{i,k}^t=\mathbf{F}(x_i^t|\hat{y}_i^t$$=$$k)\}_{i=1}^{n_b}| k\in\mathcal{C}'\}$. Correspondingly, we sample $n_b$ features from surrogate source distributions for each $k\in\mathcal{C}'$ to construct the source mini-batch $\{\{f_{j,k}^{sur}\sim\mathcal{N}_k^{sur}\}_{j=1}^{n_b}| k\in\mathcal{C}'\}$. Therefore, for any two class $k_1, k_2\in\mathcal{C}'$, a class-conditioned version of MMD that measures discrepancy between surrogate source distribution and target distribution is defined as
\begin{align}
\mathcal{L}_{\text{MMD}}^{k_1,k_2} =   \notag &\sum\limits_{i=1}^{n_b}\sum\limits_{j=1}^{n_b} \frac{\mathds{k}(f_{i,k_1}^{sur},f_{j,k_1}^{sur})}{n_b\cdot n_b} +\sum\limits_{i=1}^{n_b}\sum\limits_{j=1}^{n_b}
\frac{\mathds{k}(f_{i,k_2}^t,f_{j,k_2}^t)}{n_b\cdot n_b} \\
 -2
&\sum\limits_{i=1}^{n_b}\sum\limits_{j=1}^{n_b}
\frac{\mathds{k}(f_{i,k_1}^{sur},f_{j,k_2}^t)}{n_b\cdot n_b},
\vspace{-0.3cm}
\end{align}
where $\mathds{k}(\cdot,\cdot)$ is kernel functions that embeds feature representations in Reproducing Kernel Hilbert Space (RKHS). Utilizing the data in both source batch and target batch, the CDD loss is calculated as
\vspace{-0.2cm}
\begin{equation}
\mathcal{L}_{\text{CDD}}=\frac{~\sum\limits_{k\in\mathcal{C}'}\mathcal{L}_{\text{MMD}}^{k,k} }{|\mathcal{C}'|} - \frac{~\sum\limits_{k_1\in\mathcal{C}'}\sum\limits_{k_2\in\mathcal{C}'}^{k_1\neq k_2}\mathcal{L}_{\text{MMD}}^{k_1,k_2}}{|\mathcal{C}'|(|\mathcal{C}'|-1)},
\label{cddloss}
\vspace{-0.2cm}
\end{equation}
in which the first term represents intra-class domain discrepancy to be diminished and the second represents inter-class domain discrepancy to be enlarged. By explicitly treating data from different classes as negative sample pairs, CDD loss facilitates intra-class compactness and inter-class separability, which is beneficial to learning discriminative target features.

\cref{alg1} shows the overall training process of our proposed SFDA method within one epoch. As the adaptation proceeds, target features are driven closer and closer to approach anchors and the statistics of target features will change constantly. Therefore, we perform both pseudo-labeling method in \cref{sec:pseudolabel} and SDE method in \cref{sec:SSDE} at the beginning of every epoch to dynamically re-estimate the surrogate source distribution $\mathcal{N}_k^{sur}$. 

\section{Experiments}
\vspace{-0.15cm}
In this section, we first validate the effectiveness of the proposed SFDA-DE method based on three benchmarks. Then we conduct extensive experiments on hyper-parameter selection, ablation study, visualization, \etc.

\begin{algorithm}[t]
\caption{\mbox{SFDA training process within one epoch}}
\label{alg1}
\KwIn{unlabeled target images
$\{x_i^t\}_{i=1}^{n_t}$, label set $\mathcal{C}$, pretrained feature extractor $\mathbf{F}$, frozen classifier $\mathbf{G}$, confidence threshold $\tau$, number of iterations $t$.}
\BlankLine
Initialize the cluster center with source anchors $\mathbf{w}_k^G$ learned by $\mathbf{G}$ for each $k\in\mathcal{C}=\{1,2,\cdots,K\}$;\\
Apply spherical k-means on target features and construct confident pseudo-labeled set $\mathcal{D}_t'$ with $\tau$;\\
Perform SDE to derive K surrogate source distributions $\mathcal{N}_k^{sur} (\hat{\mu}_k^s, \hat{\Sigma}_k^s)$ according to \cref{sur_dis};\\
\For{$i={1,2,~...~,t~}$}
{
Sample target mini-batch $\{\{f_{i,k}^t=\mathbf{F}(x_{i,k}^t)\}_{i=1}^{n_b}| k\in\mathcal{C}'\}$;\\
Sample source mini-batch $\{\{f_{j,k}^{sur}\sim\mathcal{N}_k^{sur}\}_{j=1}^{n_b}| k\in\mathcal{C}'\}$;\\
Compute CDD loss according to \cref{cddloss};\\
Do backward and update weights of $\mathbf{F}$.\\
}
\end{algorithm}
\begin{table}[h]
\caption{Classification accuracy (\%) on Office-31 dataset for source-free domain adaptation (ResNet-50). Our method achieves state-of-the-art performance on A$\rightarrow$D and A$\rightarrow$W tasks. Best results under SFDA setting are shown in bold font.}
\vspace{-0.3cm}
\newcommand{\tabincell}[2]{\begin{tabular}{@{}#1@{}}#2\end{tabular}}
\centering
\resizebox{\linewidth}{!}{
\begin{tabular}{c|c|cccccc|c}
\toprule
Method & \tabincell{c}{source\\-free} & 
A$\rightarrow$D & A$\rightarrow$W & D$\rightarrow$A & D$\rightarrow$W & W$\rightarrow$A & W$\rightarrow$D & Avg. \\
\midrule
MDD\cite{zhang2019bridging} & $\times$ & 93.5 & 94.5 & 74.6 & 98.4 & 72.2 & 100.0 & 88.9\\
GVB-GD\cite{GVB} & $\times$ & 95.0 & 94.8 & 73.4 & 98.7 & 73.7 & 100 & 89.3\\
MCC\cite{MCC} & $\times$ & 95.6 & 95.4 & 72.6 & 98.6 & 73.9 & 100 & 89.4\\
GSDA\cite{GSDA} & $\times$ & 94.8 & 95.7 & 73.5 & 99.1 & 74.9 & 100 & 89.7\\
CAN\cite{CAN} & $\times$ & 95.0 & 94.5 & 78.0 & 99.1 & 77.0 & 99.8 & 90.6\\
SRDC\cite{SRDC} & $\times$ & 95.8 &  95.7 & 76.7 & 99.2 & 77.1 & 100 & 90.8\\
\midrule
SFDA\cite{kim2021domain} & \checkmark & 92.2 & 91.1 & 71.0 & 98.2 & 71.2 & 99.5 & 87.2\\
SHOT\cite{SHOT}  & \checkmark & 94.0 & 90.1 & 74.7 & 98.4 & 74.3 & 99.9 & 88.6\\
3C-GAN\cite{3CGAN}& \checkmark & 92.7 & 93.7 & 75.3 & 98.5 & \textbf{77.8} & 99.8 & 89.6\\
A$^2$Net\cite{A2Net} & \checkmark & 94.5 & 94.0 & \textbf{76.7} & \textbf{99.2} & 76.1 & \textbf{100} & \textbf{90.1}\\

SFDA-DE~(ours) & \checkmark & \textbf{96.0} & \textbf{94.2} & 76.6 & 98.5 & 75.5 & 99.8 & \textbf{90.1}\\
\bottomrule
\end{tabular}
}
\label{office31result}
\vspace{-0.2cm}
\end{table}

\begin{table*}[t]
\caption{Classification accuracy (\%) on Office-Home dateset for source-free domain adaptation (ResNet-50). Our method achieves state-of-the-art performance. Best results under SFDA setting are shown in bold font.}
\vspace{-0.3cm}

\centering
\resizebox{\linewidth}{!}{
\begin{tabular}{c|c|cccccccccccc|c}
\toprule
Method & source-free & 
Ar$\rightarrow$Cl & Ar$\rightarrow$Pr & Ar$\rightarrow$Rw & Cl$\rightarrow$Ar & Cl$\rightarrow$Pr & Cl$\rightarrow$Rw & Pr$\rightarrow$Ar & Pr$\rightarrow$Cl & Pr$\rightarrow$Rw & Rw$\rightarrow$Ar & Rw$\rightarrow$Cl & Rw$\rightarrow$Pr & Avg. \\
\midrule

GSDA\cite{GSDA} & $\times$ & 61.3 & 76.1 & 79.4 & 65.4 & 73.3 & 74.3 & 65.0 & 53.2 & 80.0 & 72.2 & 60.6 & 83.1 & 70.3\\
GVB-GD\cite{GVB} & $\times$ & 57.0 & 74.7 & 79.8 & 64.6 & 74.1 & 74.6 & 65.2 & 55.1 & 81.0 & 74.6 & 59.7 & 84.3 & 70.4\\
RSDA\cite{RSDA} & $\times$ & 53.2 & 77.7 & 81.3 & 66.4 & 74.0 & 76.5 & 67.9 & 53.0 & 82.0 & 75.8 & 57.8 & 85.4 & 70.9 \\
TSA\cite{li2021transferable} & $\times$ & 57.6 & 75.8 & 80.7 & 64.3 & 76.3 & 75.1 & 66.7 & 55.7 & 81.2 & 75.7 & 61.9 & 83.8 & 71.2\\
SRDC\cite{SRDC} & $\times$ & 52.3 & 76.3 & 81.0 & 69.5 & 76.2 & 78.0 & 68.7 & 53.8 & 81.7 & 76.3 & 57.1 & 85.0 & 71.3 \\
FixBi\cite{FixBi} & $\times$ & 58.1 & 77.3 & 80.4 & 67.7 & 79.5 & 78.1 & 65.8 & 57.9 & 81.7 & 76.4 & 62.9 & 86.7 & 72.7 \\
\midrule
SFDA\cite{kim2021domain} & \checkmark & 48.4 & 73.4 & 76.9 & 64.3 & 69.8 & 71.7 & 62.7 & 45.3 & 76.6 & 69.8 & 50.5 & 79.0 & 65.7\\
G-SFDA\cite{G-SFDA} & \checkmark & 57.9 & 78.6 & 81.0 & 66.7 & 77.2 & 77.2 & 65.6 & 56.0 & 82.2 & 72.0 & 57.8 & 83.4 & 71.3\\
SHOT\cite{SHOT} & \checkmark & 57.1 & 78.1 & 81.5 & 68.0 & 78.2 & 78.1 & 67.4 & 54.9 & 82.2 & 73.3 & 58.8 & 84.3 & 71.8\\
A$^2$Net\cite{A2Net} & \checkmark & 58.4 & 79.0 & \textbf{82.4} & 67.5 & \textbf{79.3} & 78.9 & \textbf{68.0} & 56.2 & \textbf{82.9} & \textbf{74.1} & 60.5 & 85.0 & 72.8\\

SFDA-DE~(ours) & \checkmark & \textbf{59.7} & \textbf{79.5} & \textbf{82.4} & \textbf{69.7} & 78.6 & \textbf{79.2} & 66.1 & \textbf{57.2} & 82.6 & 73.9 & \textbf{60.8} & \textbf{85.5} & \textbf{72.9}\\
\bottomrule
\end{tabular}}
\label{officehomeresult}
\end{table*}


\begin{table*}[t]
\caption{Per-class accuracy and mean accuracy (\%) on VisDA-2017 dateset for source-free domain adaptation (ResNet-101). Our method achieves state-of-the-art performance. Best results under SFDA setting are shown in bold font.}
\vspace{-0.3cm}
\footnotesize
\centering
\begin{tabular}{c|c|cccccccccccc|c}
\toprule
Method & source-free & plane &  bike &  bus &  car &  horse &  knife & mcycle &  person &  plant &  sktbrd &  train &  truck &  Avg. \\
\midrule
SFAN\cite{SFAN} & $\times$  &  93.6 & 61.3 & 84.1 & 70.6 & 94.1 & 79.0 & 91.8 & 79.6 & 89.9 & 55.6 & 89.0 & 24.4 & 76.1\\
SWD\cite{lee2019sliced} & $\times$ & 90.8 & 82.5 & 81.7 & 70.5 & 91.7 & 69.5 & 86.3 & 77.5 & 87.4 & 63.6 & 85.6 & 29.2 & 76.4 \\
MCC\cite{MCC} & $\times$ & 88.7 & 80.3 & 80.5 & 71.5 & 90.1 & 93.2 & 85.0 & 71.6 & 89.4 & 73.8 & 85.0 & 36.9 & 78.8\\
STAR\cite{STAR} & $\times$ & 95.0 & 84.0 & 84.6 & 73.0 & 91.6 & 91.8 & 85.9 & 78.4 & 94.4 & 84.7 & 87.0 & 42.2 & 82.7 \\
RWOT\cite{RWOT} & $\times$ & 95.1 & 80.3 & 83.7 & 90.0 & 92.4 & 68.0 & 92.5 & 82.2 & 87.9 & 78.4 & 90.4 & 68.2 & 84.0 \\
SE\cite{SE} & $\times$ & 95.9 & 87.4 & 85.2 & 58.6 & 96.2 & 95.7 & 90.6 & 80.0 & 94.8 & 90.8 & 88.4 & 47.9 & 84.3\\

\midrule
SFDA\cite{kim2021domain} & \checkmark &86.9 & 81.7 & 84.6 & 63.9 & 93.1 & 91.4 & 86.6 & 71.9 & 84.5 & 58.2 & 74.5 & 42.7 & 76.7\\
3C-GAN\cite{3CGAN} & \checkmark & 94.8 & 73.4 & 68.8 & 74.8 & 93.1 & 95.4 & 88.6 & 84.7 & 89.1 & 84.7 & 83.5 & 48.1 & 81.6\\
SHOT\cite{SHOT} & \checkmark & 94.3 & 88.5 & 80.1 & 57.3 & 93.1 & 94.9 & 80.7 & 80.3 & 91.5 & 89.1 & 86.3 & 58.2 & 82.9\\
A$^2$Net\cite{A2Net} & \checkmark & 94.0 & 87.8 & 85.6 & 66.8 & 93.7 & 95.1 & 85.8 & 81.2 & 91.6 & 88.2 & 86.5 & 56.0 & 84.3\\
G-SFDA\cite{G-SFDA} & \checkmark & \textbf{96.1} & 88.3 & \textbf{85.5} & \textbf{74.1} & \textbf{97.1} & 95.4 & \textbf{89.5} & 79.4 & 95.4 & 92.9 & \textbf{89.1} & 42.6 & 85.4\\
SFDA-DE~(ours) & \checkmark & 95.3 & \textbf{91.2} & 77.5 & 72.1 & 95.7 & \textbf{97.8} & 85.5 & \textbf{86.1} & \textbf{95.5} & \textbf{93.0} & 86.3 & \textbf{61.6} & \textbf{86.5}\\
\bottomrule
\end{tabular}
\label{visdaresult}
\end{table*}

\vspace{-0.2cm}
\subsection{Experimental settings}
\paragraph{Office-31 dataset.}
Office-31~\cite{office31} is a small-scale benchmark with 3 domains, \textbf{A}mazon (2,817), \textbf{D}SLR (498) and \textbf{W}ebcam (795) . There are totally 4,110 images belonging to 31 categories collected from real world scenarios.

\newcommand{\paravspace}{-0.5}
\vspace{\paravspace cm}
\paragraph{Office-Home dataset.}
Office-Home~\cite{venkateswara2017deep} is a complex benchmark comprised of four visually-dissimilar domains: \textbf{Ar}tistic images, \textbf{Cl}ipart images, \textbf{Pr}oduct images, and \textbf{R}eal-\textbf{w}orld images. This dataset contains 12 transfer tasks and a total number of 15,500 images from 65 classes.

\vspace{\paravspace cm}
\paragraph{VisDA-2017 dataset.}
VisDA-2017~\cite{peng2017visda} is a large-scale synthetic-to-real dataset with 12 classes in both domain. The synthetic domain contains 150K rendered 3D images with various poses and lighting conditions. The corresponding real domain contains about 55K real-world images.

\vspace{\paravspace cm}
\paragraph{Pretraining on source domain.}
We use momentum SGD optimizer with exponential decay learning rate schedule $\eta = \eta_0 (1 + \alpha\cdot i)^{-\beta}$, where $\eta_0$ is the initial learning rate and $i$ is the training steps. Weight decay is set to 5e-4 and momentum is set to 0.9.
For Office-31 and Office-Home dataset, we employ ResNet-50~\cite{he2016deep} as our feature extractor $\mathbf{F}$ and a single fully-connected layer as classifier $\mathbf{G}$. We set $\eta_0=0.001$, $\alpha=0.001$ and $\beta=0.75$. The model is trained for 50 epochs on all source domains. For VisDA-2017 dataset, we employ ResNet-101 as the feature extractor $\mathbf{F}$ and train it for 500 steps on source domain. We set $\eta_0=0.001$, $\alpha=0.0005$ and $\beta=2.25$. For all 3 datasets, the learning rate of $\mathbf{G}$ is set to be 10 times bigger than $\mathbf{F}$ and the batch size is set to 64 for all domains. The source dataset is randomly split into a training set accounting for 90\% and a validation set accounting for 10\% in order to guarantee the model converges.

\vspace{\paravspace cm}
\paragraph{SFDA implementation detail.}
We follow the standard SFDA setups adopted by~\cite{SHOT,A2Net}. We use all weights of the pretrained model as initialization and freeze all anchors $\mathbf{w}_k^G$ in classifier $\mathbf{G}$ during SFDA stage. For Office31 and Office-Home dataset, we use the same optimization setting and learning rate schedule as the aforementioned pretraining stage. We empirically set $\tau=0.6$, $\gamma=1$, $|\mathcal{C}'|=12$ and $n_b=3$. For VisDA-2017 dataset, we use the same optimization setting and learning rate schedule as the pretraining stage but set the initial learning rate $\eta_0$ to be 1e-4 for all convolutional layers and 1e-3 for all BatchNorm layers. We empirically set $\tau=0.078$, $\gamma=2$, $|\mathcal{C}'|=6$ and $n_b=10$. Selection of hyper-parameters will be studied in \cref{sec:abl}. All results reported below are the average of 3 independent runs and we manually set the random seed to guarantee reproducibility. All experiments are conducted with PyTorch and MindSpore~\cite{mindspore} on NVIDIA 1080Ti GPUs.

\begin{figure*}[t]
\centering
\begin{subfigure}{0.252\linewidth}
\centering
\includegraphics[width=0.99\linewidth]{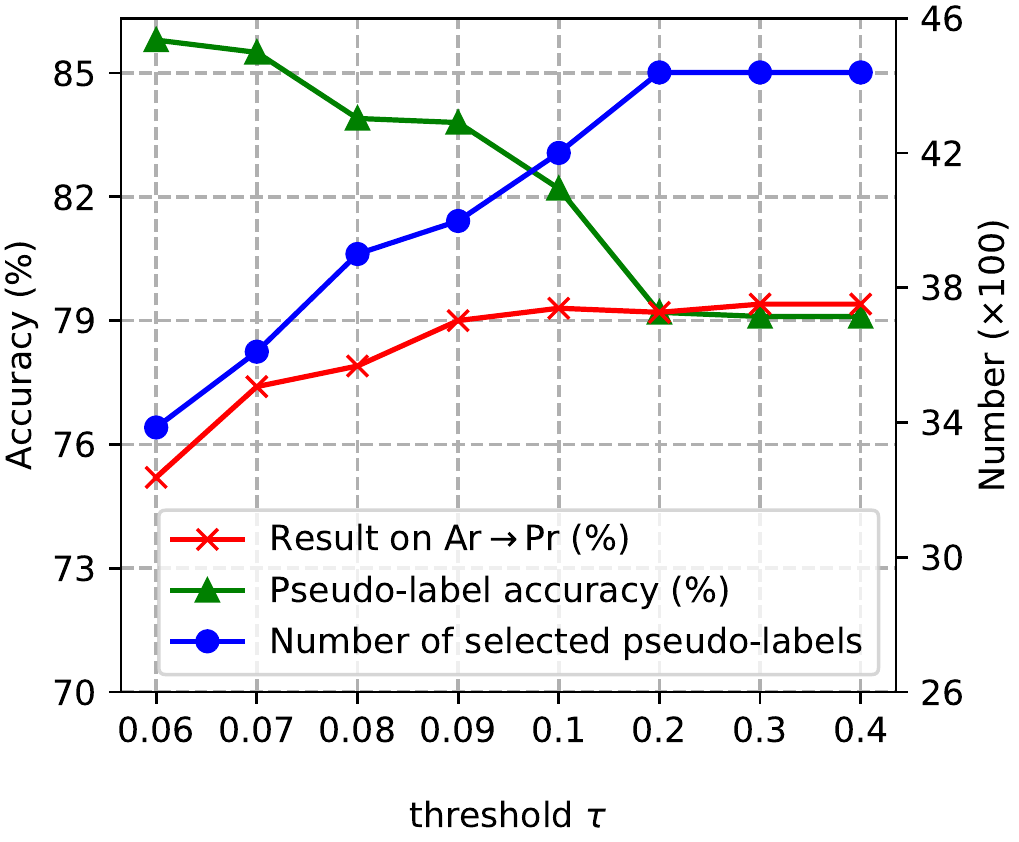}
\caption{Ar$\rightarrow$Pr}
\label{tau-a}
\end{subfigure}
\begin{subfigure}{0.252\linewidth}
\centering
\includegraphics[width=0.99\linewidth]{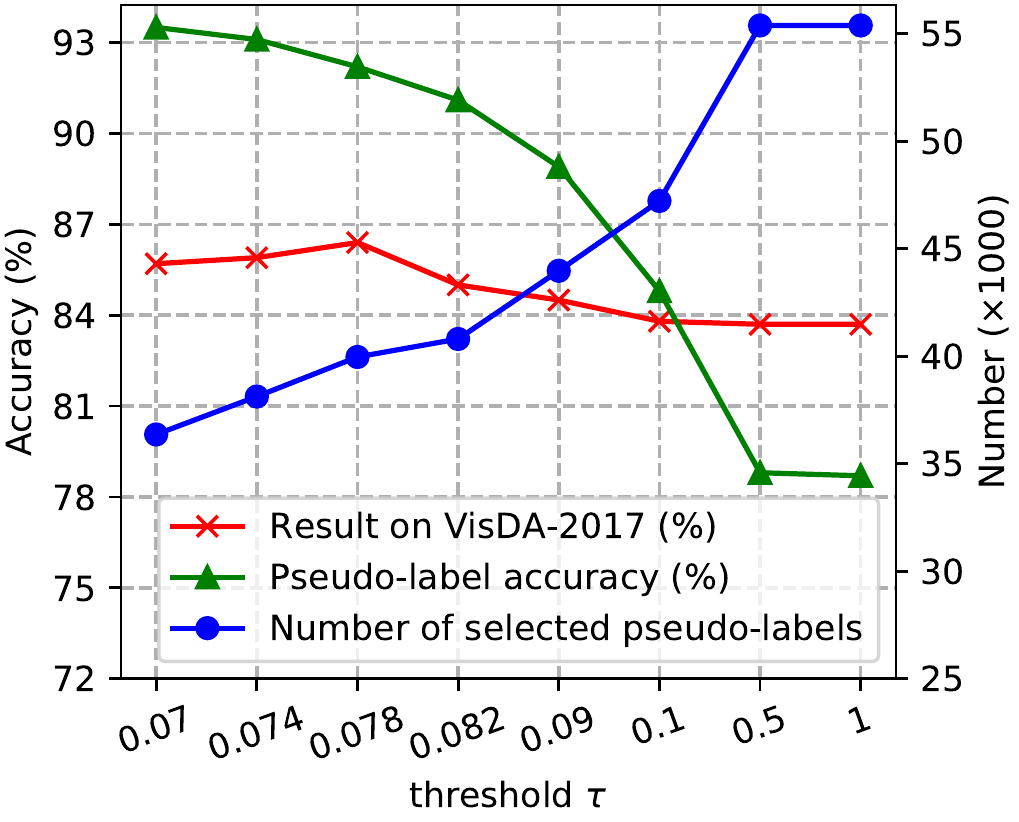}
\caption{VisDA-2017}
\label{tau-b}
\end{subfigure}
\begin{subfigure}{0.24\linewidth}
\centering
\includegraphics[width=0.9\linewidth]{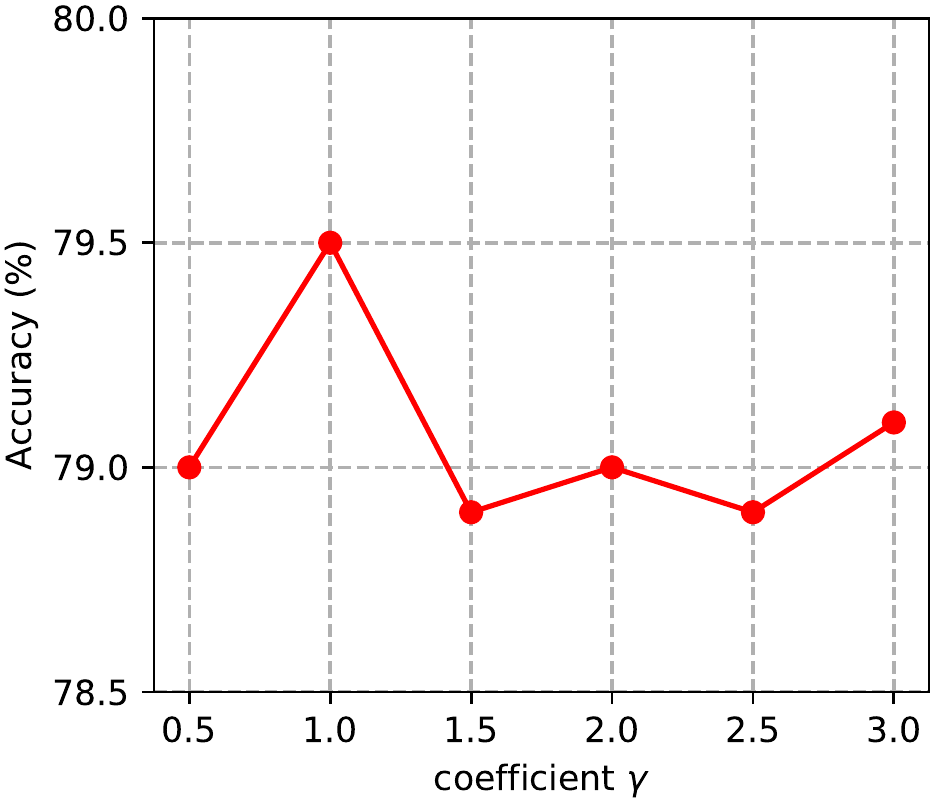}
\caption{Ar$\rightarrow$Pr}
\label{gamma-ab-1}
\end{subfigure}
\begin{subfigure}{0.24\linewidth}
\centering
\includegraphics[width=0.9\linewidth]{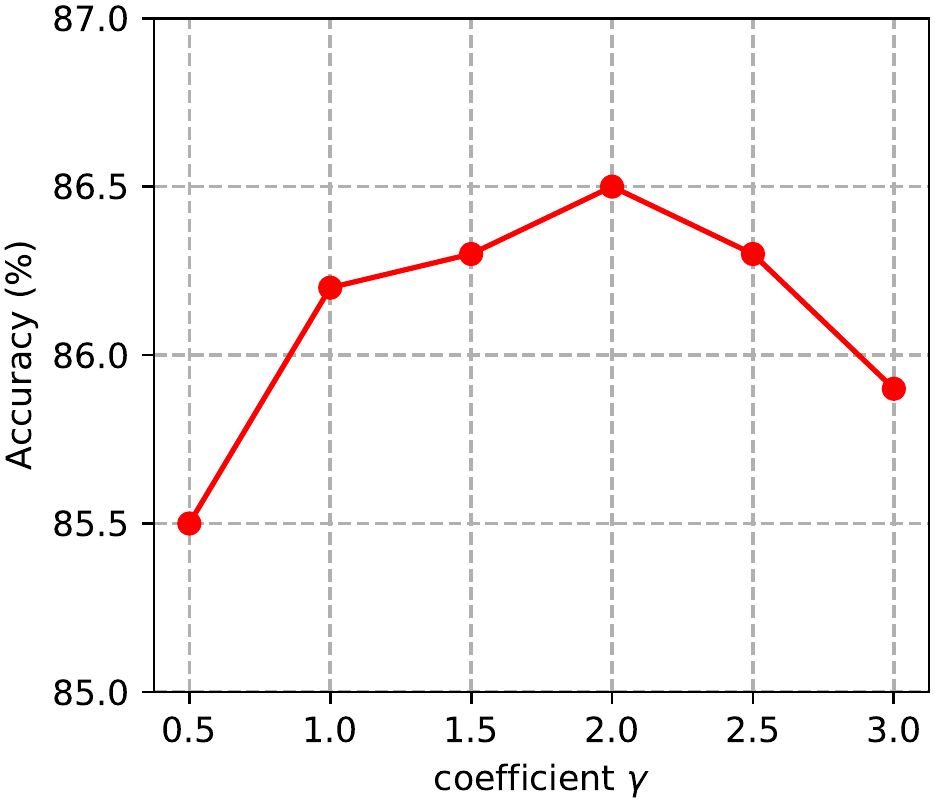}
\caption{VisDA-2017}
\label{gamma-ab-2}
\end{subfigure}
\vspace{-0.4cm}
\caption{Analysis on hyper-parameter sensitivity. (a) and (b) Sensitivity of pseudo-labels and model performance w.r.t. $\tau$. (c) and (d) Sensitivity of model performance w.r.t. $\gamma$.}
\vspace{-0.3cm}
\label{fig:ab:tau}
\end{figure*}

\vspace{-0.1cm}
\subsection{Experimental results}
\vspace{-0.1cm}
\cref{office31result,officehomeresult,visdaresult} demonstrate the experimental results of several recent SFDA methods and traditional DA methods. Best results among SFDA methods are shown in bold font. We achieve state-of-the-art performance on Office-Home (72.9\%) and VisDA-2017 (86.5\%). As the scale of dataset gets larger, our method performs increasingly better.

\cref{office31result} shows the adaptation results on Office-31 dataset. Our method has the same best result (90.1\%) as A$^2$Net~\cite{A2Net} and is comparable to some traditional domain adaptation algorithms which require source data. Unlike Office-Home and VisDA-2017, Office-31 is a small-scale dataset whose image number of each class is around 40 on average. Therefore, it is hard for our method to accurately estimate the source distributions from statistics of target data. Yet we still achieve the best results on average and on 2 of 6 tasks.

\vspace{-0.03cm}
\cref{officehomeresult} shows the results on Office-Home benchmark, in which our method achieves state-of-the-art average performance (72.9\%) and performs the best on 7 of 12 transfer tasks among all the SFDA methods. Our method is even superior to some of the traditional domain adaptation methods which require source data. This dataset is larger in scale than Office-31 and thus is able to provide adequate target data to estimate source distributions more accurately.

\cref{visdaresult} shows the per-class and average accuracy on VisDA-2017 benchmark. Our method achieves state-of-the-art performance among all SFDA methods and is higher than the second best A$^2$Net~\cite{A2Net} by a margin of 1.1\%. Despite the huge domain gap between the source domain (Synthetic) and the target (Real), our method still achieves 86.5\% average accuracy due to the vast number of target images ($\sim$55K) for estimating the source distributions, which is the key to our success. Statistics derived from sufficient of data can better reflect the real distribution.

\subsection{Ablation studies}
\label{sec:abl}
\paragraph{Confidence threshold $\tau$.} The precision of the estimation of class-conditioned source distributions relies on the correctness of target pseudo-labels included by $\mathcal{D}_t'$ in \cref{confidentpseido}. \cref{tau-a,tau-b} shows the sensitivity analysis on model performance, pseudo-label accuracy of $\mathcal{D}_t'$ and the number of data included by $\mathcal{D}_t'$~w.r.t. confidence threshold $\tau$. Specifically, a small threshold $\tau$ would reject more incorrectly labeled data but the total number of data in $\mathcal{D}_t'$ would be reduced. Conversely, a large threshold will enlarge the scale of $\mathcal{D}_t'$ but introduce more false labels. Therefore, $\tau$ needs to be selected carefully. As shown in \cref{tau-a}, for Ar$\rightarrow$Pr task in Office-Home dataset, despite the drop in pseudo-label accuracy caused by increasing $\tau$, the performance of our method keeps improving in synchronization with the number of target data included by $\mathcal{D}_t'$. We conjecture that having sufficient pseudo-labeled data is more important than the accuracy of pseudo-labels to the estimation of source distributions for small-scale dataset. So we set $\tau$~=~0.6 for both Office-31 and Office-Home to allow more pseudo-labels. However for VisDA-2017 dataset, as shown in \cref{tau-b}, the best performance is obtained when $\tau$~=~0.078. Since VisDA is a large-scale dataset, a small $\tau$ can guarantee both the accuracy of pseudo-labels and the number of selected confident data simultaneously.

\begin{figure*}[t]
\centering
\begin{subfigure}{0.27\linewidth}
\centering
\includegraphics[width=0.80\linewidth]{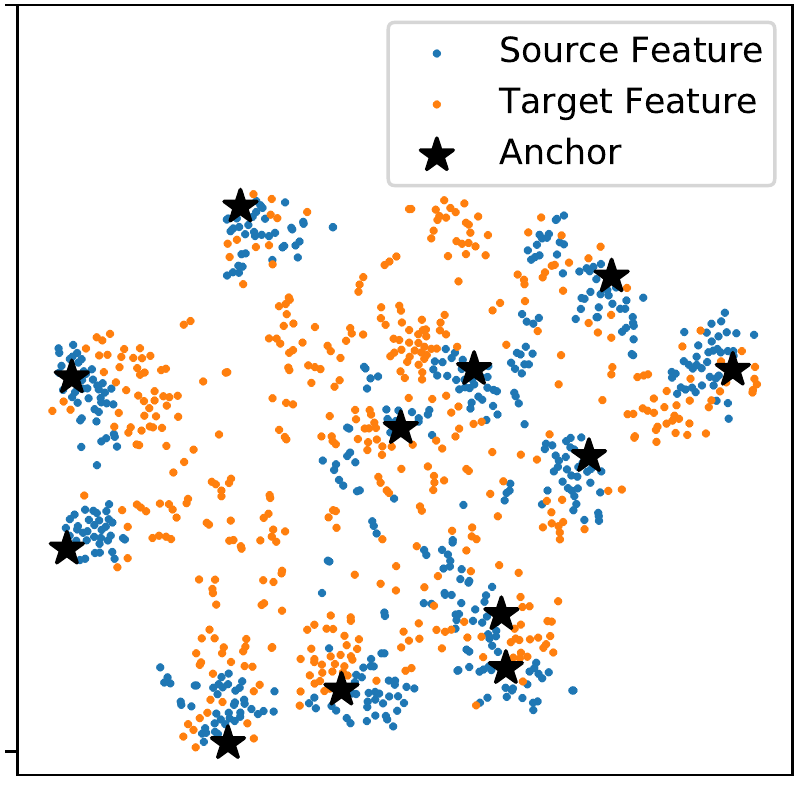}
\vspace{-0.1cm}
\caption{Before Adaptation}
\label{vis_ana-a}
\end{subfigure}
\begin{subfigure}{0.27\linewidth}
\centering
\includegraphics[width=0.80\linewidth]{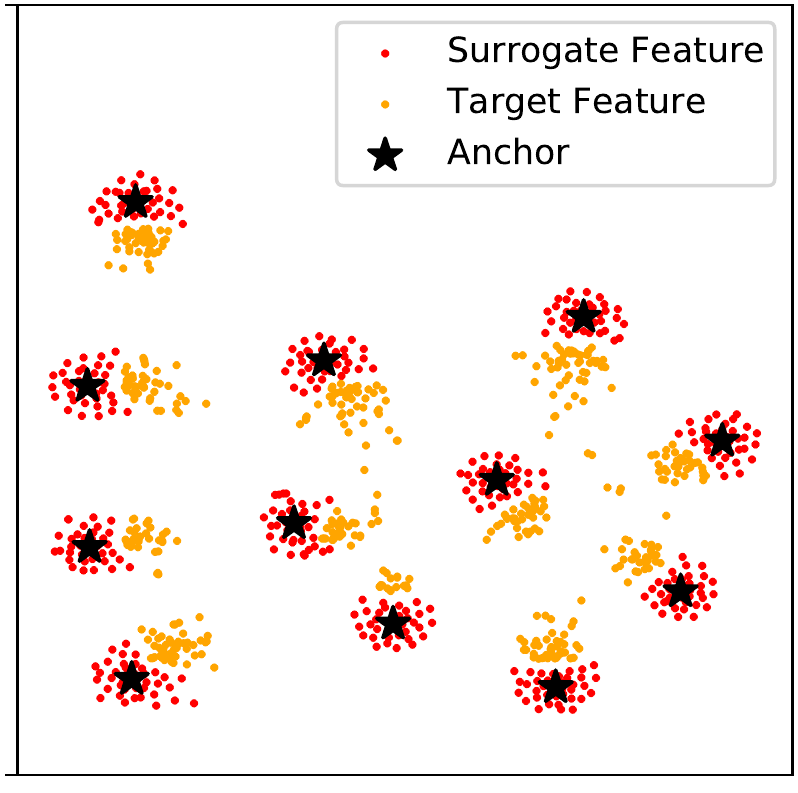}
\vspace{-0.1cm}
\caption{After Adaptation}
\label{vis_ana-b}
\end{subfigure}
\begin{subfigure}{0.328\linewidth}
\centering
\includegraphics[width=0.93\linewidth]{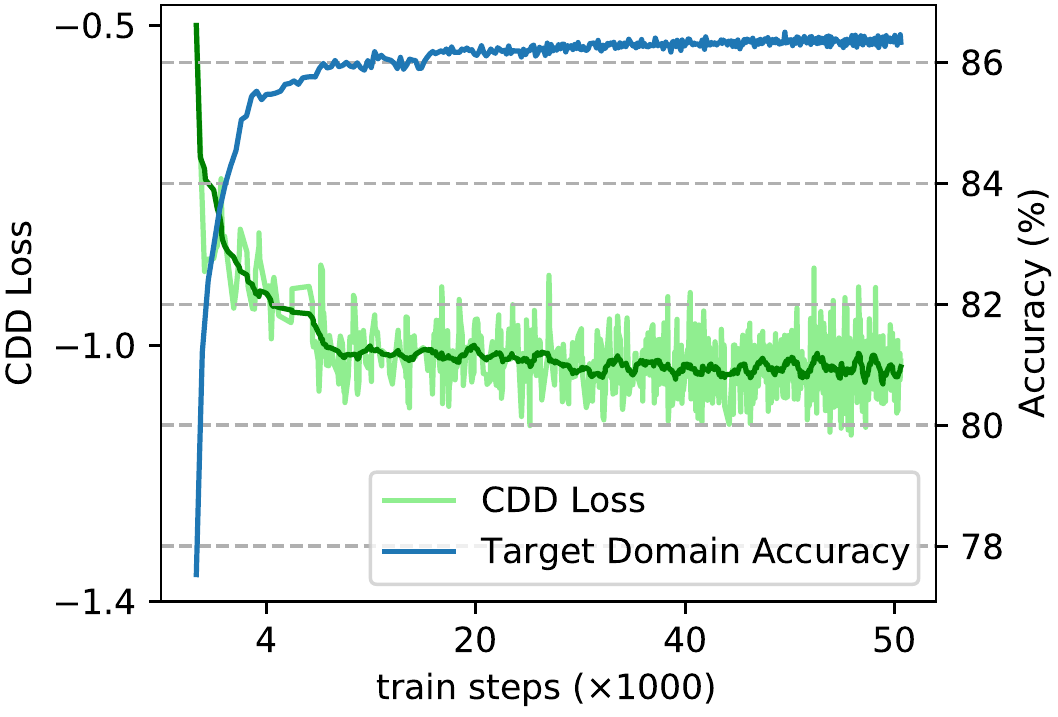}
\caption{Training Curves}
\label{vis_ana-c}
\end{subfigure}
\vspace{-0.3cm}
\caption{Visualization on VisDA-2017 dataset. (a) T-SNE visualization of source features and target features before SFDA. (b) T-SNE visualization of surrogate features and target features after SFDA. (c) Curves of CDD loss and model performance on target domain.}
\vspace{-0.2cm}
\label{fig:vis-ana}
\end{figure*}

\begin{figure}[h]
\centering
\begin{subfigure}{0.49\linewidth}
\includegraphics[width=0.99\linewidth]{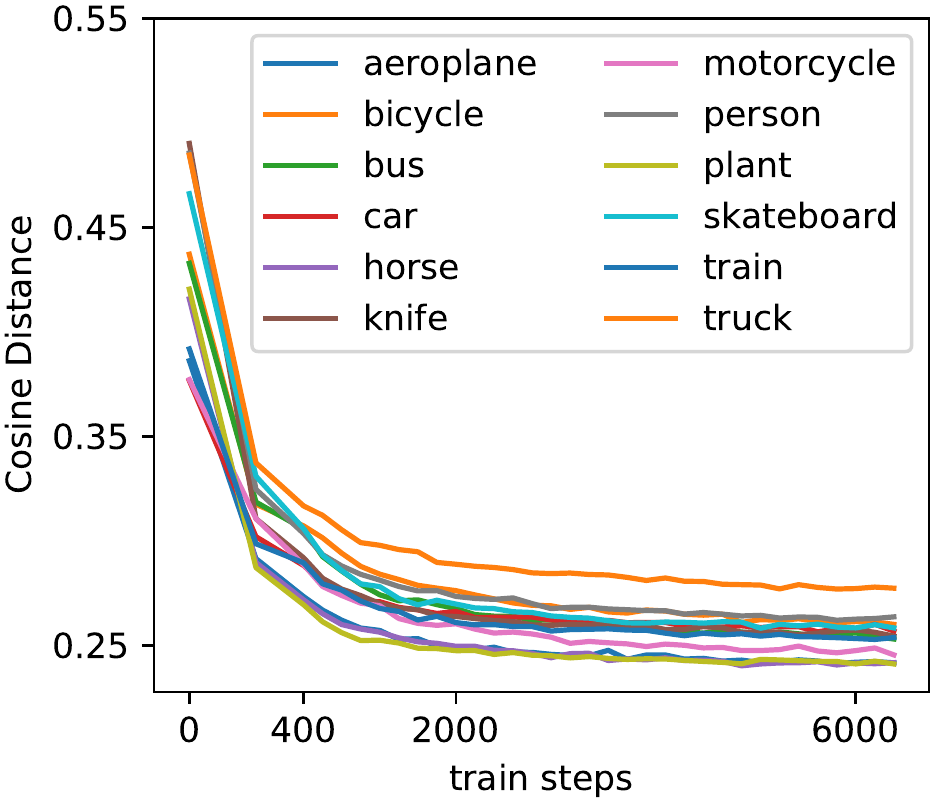}
\vspace{-0.5cm}
\caption{}
\label{cdist-a}
\end{subfigure}
\begin{subfigure}{0.49\linewidth}
\includegraphics[width=0.99\linewidth]{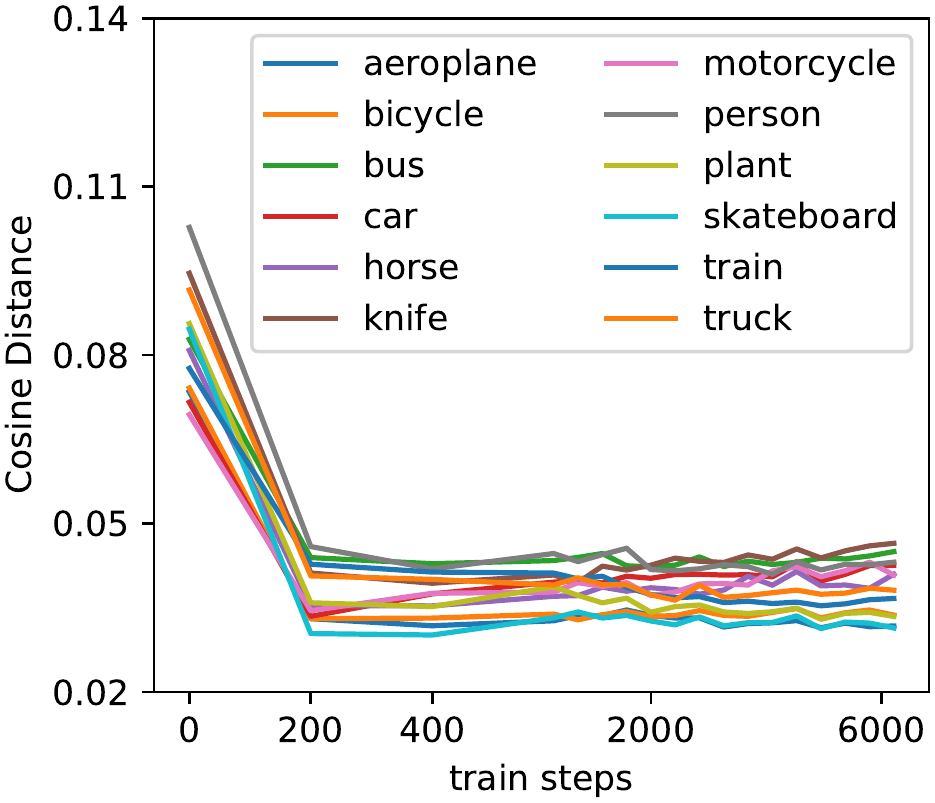}
\vspace{-0.5cm}
\caption{}
\label{cdist-b}
\end{subfigure}
\vspace{-0.3cm}
\caption{(a) Cosine distance between the centers of target features $\bar{f_k^t}$ and source anchors $\mathbf{w}_k^G$ for each class. (b) Cosine distance between target covariance $\Sigma_k^t$ and corresponding source covariance $\Sigma_k^s$ for each class. }
\vspace{-0.3cm}
\label{fig:cosdist}
\end{figure}

\vspace{\paravspace cm}
\paragraph{Covariance coefficient $\gamma$.}
\cref{gamma-ab-1,gamma-ab-2} shows the experimental results with different $\gamma\in\{0.5,1,1.5,2,2.5,3\}$ on Ar$\rightarrow$Pr task of Office-Home dataset and on VisDA-2017 dataset, respectively. Larger covariance matrix leads to more flexible feature activations. Thus the value of $\gamma$ in \cref{sigmaestim} controls the semantic diversity of features sampled from the surrogate source distribution. By expanding the sampling range, features far from anchors can be sampled. \cref{gamma-ab-2} shows that the performance on VisDA-2017 is improved by a margin of 0.2\% when $\gamma$~=~2. However, an inappropriate value of $\gamma$ may lead to a sub-optimal solution.

\vspace{\paravspace cm}
\paragraph{Estimation of the source mean.}
To verify the effectiveness of the estimator $\hat{\mu}_k^s$ in \cref{muestim}, we use several variants to replace our estimation. If we directly treat the intra-class feature mean derived from confident target data in $\mathcal{D}_t'$ as the mean of surrogate source distribution $\hat{\mu}_k^s$$=$$\bar{f_k^t}$$=$$\frac{\sum_{i}f_{i,k}^t}{\sum_{x_i^t \in \mathcal{D}_t'}\mathds{1}(\hat{y}_i^t=k)}$, as shown in \cref{variants_mu}, the performance of our method decreases. Especially on VisDA-2017, the performance drops by 17.7\%. On the other hand, if we directly use the anchors as estimated mean $\hat{\mu}_k^s=\mathbf{w}_k^G$, the performance becomes even worse. This suggests that information of source anchors and information of target features complement each other. In addition, \textit{update once} in \cref{variants_mu} means we only update both $\hat{\mu}_k^s$ and $\hat{\Sigma}_k^s$ for once at the very beginning of the whole SFDA training process, which leads to a sub-optimal result.

\begin{table}[t]
\caption{Performance with different $\hat{\mu}_k^s$ on Ar$\rightarrow$Cl, Ar$\rightarrow$Pr, Ar$\rightarrow$Rw tasks (Office-Home) and VisDA-2017 dataset.}
\vspace{-0.3cm}
\centering
\footnotesize
\begin{tabular}{c|cccc|c}
\toprule
estimator & Ar$\rightarrow$Cl & Ar$\rightarrow$Pr & Ar$\rightarrow$Rw & Avg. & VisDA\\
\midrule
$\hat{\mu}_k^s=\bar{f_k^t}$ & 59.2 & 78.0 & 80.2 & 72.5 & 68.8\\
$\hat{\mu}_k^s=\mathbf{w}_k^G$ & 47.1 & 68.8 & 76.2 & 64.0 & 64.1\\
update once & 55.7 & 79.2 & 81.1 & 72.0 & 79.7\\
Ours & \textbf{59.7} & \textbf{79.5} & \textbf{82.4} & \textbf{73.9} & \textbf{86.5}\\
\bottomrule
\end{tabular}
\label{variants_mu}
\end{table}
\begin{table}[t]
\caption{Comparing with maximum probability-based pseudo-labeling method on Ar$\rightarrow$Cl, Ar$\rightarrow$Pr, Ar$\rightarrow$Rw tasks (Office-Home) and VisDA-2017 dataset.}
\vspace{-0.3cm}

\centering
\footnotesize
\begin{tabular}{c|cccc|c}
\toprule
$\tau'$ & Ar$\rightarrow$Cl & Ar$\rightarrow$Pr & Ar$\rightarrow$Rw & Avg. & VisDA\\
\midrule
0.975 & 48.8 & 74.1 & 77.2 & 66.7 & 85.7\\
0.950 & 55.8 & 76.3 & 79.6 & 70.6 & 85.8\\
0.925 & 58.1 & 76.4 & 80.0 & 71.5 & 85.5\\
0.900 & 57.8 & 78.3 & 81.4 & 72.5 & 85.6\\
0.875 & 58.4 & 78.9 & 82.3 & 73.2 & 85.5\\
0.850 & 59.0 & 78.8 & 81.6 & 73.1 & 85.3\\
Ours & \textbf{59.7} & \textbf{79.5} & \textbf{82.4} & \textbf{73.9} & \textbf{86.5}\\
\bottomrule
\end{tabular}
\label{pmaxplabel}
\end{table}

\vspace{\paravspace cm}
\paragraph{Robustness of pseudo-labeling strategy.}
Obtaining robust pseudo-labels is important to the following SDE process, since high-quality pseudo-labels can provide accurate estimation for the mean and covariance of each distribution. If pseudo-labels are corrupted, the estimated distribution would be diverged from the real distribution, which makes the sampled surrogate features unable to represent the real source features of a certain class. To validate the robustness of our anchor-based spherical k-means clustering pseudo-labeling method, we conduct experiments and show the results in \cref{pmaxplabel}. Instead, we use a maximum probability-based strategy to assign pseudo-labels: $\hat{y}_i^t = \mathop{\arg\max}_k~\sigma_k(\mathbf{G}(\mathbf{F}(x_i^t)))$, where $\sigma$ is the $K$-way softmax function that generate probabilities for each class. We also set a threshold $\tau'$ to select confident samples whose maximum probabilities are greater than $\tau'$ to construct the confident target dataset $\mathcal{D}_t'$. Multiple values of $\tau'$ are tested to guarantee a fair comparison. \cref{pmaxplabel} shows that our anchor-based clustering pseudo-labeling method outperforms maximum probability-based method on both Office-Home dataset and VisDA-2017 dataset.

\renewcommand{\paravspace}{-0.35}
\subsection{Visualization and empirical analysis}
We visualize the experimental results on VisDA-2017 dataset and analyse the proposed SFDA-DE method.

\vspace{\paravspace cm}
\paragraph{Training curves.}
\cref{vis_ana-c} shows the training curves of CDD loss and model accuracy on target domain during source-free adaptation process. Our method converges stably and shows superior performance from an early stage.

\vspace{\paravspace cm}
\paragraph{Domain shift.}
We utilize t-SNE visualization to demonstrate the distributions of feature representations in both source and target domains. As shown in \cref{vis_ana-a}, a large amount of target data (represented by orange dots) disperses in the feature space before adaptation due to severe domain shift problem while source data (represented by blue dots) gathers around the anchors and forms intra-class clusters. 

\renewcommand{\paravspace}{-0.35}

\vspace{\paravspace cm}
\paragraph{Visualization of surrogate features.} Red dots in \cref{vis_ana-b} represent the surrogate features derived from SDE with covariance multiplier $\gamma=2$, which enlarges the sampling range. These surrogates are distributed around corresponding anchors to simulate source features of each class.

\vspace{\paravspace cm}
\paragraph{Effectiveness of our method.}
After SFDA training, as shown in \cref{vis_ana-b}, target features are pulled towards corresponding anchors and merged into the surrogate feature clusters. Besides, low density area can be clearly observed in the feature space after adaptation. This suggests our SFDA-DE method can learn discriminative features for unlabeled target domain without using source data.

\vspace{\paravspace cm}
\paragraph{Calibration of distribution mean.}
In SDE, anchors are utilized to calibrate the mean of estimated source distribution according to \cref{sigmaestim}, since target features drift away from source features at the early stage of training. Therefore, target class centers $\bar{f_k^t}$$=$$\frac{\sum_{i}f_{i,k}^t}{\sum_{x_i^t \in \mathcal{D}_t'}\mathds{1}(\hat{y}_i^t=k)}$ cannot serve as a good estimator of $\mu_k^s$. As shown in \cref{cdist-a}, the distance between target feature centers and source anchors is diminished as training proceeds. Target features gradually approach the corresponding anchors of the same class, which means the calibration of $\hat{\mu}_k^s$ is effective.

\vspace{\paravspace cm}
\paragraph{Estimation bias of covariance.}
\cref{cdist-b} visualizes the classwise estimation bias of distribution covariance $\hat{\Sigma}_k^s$$=$$\Sigma_k^t$ over the ground truth source covariance $\Sigma_k^s$ w.r.t. training steps. The gap in between is mitigated in the early stage of training and is kept at a low level, which verifies our assumption made in \cref{sec:SSDE}. Thus, class-conditioned source covariance can be approximated via high-quality pseudo-labeled target data.

\section{Conclusions}
\vspace{-0.15 cm}
In this paper, we propose a novel framework named SFDA-DE to address source-free domain adaptation problem via estimating feature distributions of source domain in the absence of source data. We utilize domain knowledge preserved by source anchors to obtain high-quality pseudo-labels for target data to achieve our goal.
Sufficient experiments validate the effectiveness and superiority of our method against other strong SFDA baselines.

\section*{Acknowledgements}
This work is supported by National Natural Science Foundation of China under Grant No. 61876007.


\newpage

{\small
\bibliographystyle{ieee_fullname}
\bibliography{egbib}
}

\end{document}